\def\year{2020}\relax
\documentclass[letterpaper]{article} % DO NOT CHANGE THIS
\usepackage{aaai20}  % DO NOT CHANGE THIS
\usepackage{times}  % DO NOT CHANGE THIS
\usepackage{helvet} % DO NOT CHANGE THIS
\usepackage{courier}  % DO NOT CHANGE THIS
\usepackage[hyphens]{url}  % DO NOT CHANGE THIS
\usepackage{graphicx} % DO NOT CHANGE THIS
\usepackage{graphicx}  % DO NOT CHANGE THIS
\usepackage{amsmath,amssymb}
\usepackage{mathtools}
\usepackage{color}
\graphicspath{{figures/}}

\usepackage{algpseudocode,algorithm}

\algnewcommand\algorithmicinput{\textbf{Input:}}
\algnewcommand\Input{\item[\algorithmicinput]}
\algnewcommand\algorithmicoutput{\textbf{Output:}}
\algnewcommand\Output{\item[\algorithmicoutput]}
\algnewcommand\algorithmictier{\textbf{Step:}}
\algnewcommand\Tier{\item[\algorithmictier]}

\newcommand{\citet}[1]{\citeauthor{#1}~\shortcite{#1}}
\newcommand{\citep}{\cite}

\DeclareMathOperator*{\argmax}{\arg\!\max}

\begin{document}
%\nocopyright
%PDF Info Is REQUIRED.
% For /Author, add all authors within the parentheses, separated by commas. No accents or commands.
% For /Title, add Title in Mixed Case. No accents or commands. Retain the parentheses.
 \pdfinfo{
/Title (Hierarchical Reinforcement Learning in StarCraft II with Human Expertise in Subgoals Selection)
/Author (Xu et al.)
} %Leave this	

\setcounter{secnumdepth}{0} %May be changed to 1 or 2 if section numbers are desired.

% The file aaai20.sty is the style file for AAAI Press 
\setlength\titlebox{2.5in} % If your paper contains an overfull \vbox too high warning at the beginning of the document, use this
% command to correct it. You may not alter the value below 2.5 in
\title{Hierarchical Reinforcement Learning in StarCraft II \\with Human Expertise in Subgoals Selection}% Force line breaks with \\%Your title must be in mixed case, not sentence case. 
% That means all verbs (including short verbs like be, is, using,and go), 
% nouns, adverbs, adjectives should be capitalized, including both words in hyphenated terms, while articles, conjunctions, and prepositions are lower case unless they directly follow a colon or long dash
\author{Xinyi Xu\thanks{Corresponding author}\textsuperscript{\rm 1},
Tiancheng Huang\textsuperscript{\rm 2}\\ \Large\textbf{Pengfei Wei\textsuperscript{\rm 1}, Akshay Narayan\textsuperscript{\rm 1}, Tze-Yun Leong\textsuperscript{\rm 1}} \\ 
% All authors must be in the same font size and format. Use \Large and \textbf to achieve this result when breaking a line
%If you have multiple authors and multiple affiliations
% use superscripts in text and roman font to identify them. For example,
\textsuperscript{1}{NUS, School of Computing, Medical Computing Lab},\\
{\{xuxinyi,weipf,anarayan,leongty\}@comp.nus.edu.sg} \\
\textsuperscript{2}{NTU, School of Computer Science and Engineering},\\
thuang013@e.ntu.edu.sg 
%Note that the comma should be placed BEFORE the superscript for optimum readability
% 13 Computing Drive, NUS School of Computing \\
% Singapore, Singapore 117417 \\
% \{xuxinyi,huangtc,weipf,anarayan,leongty\}@comp.nus.edu.sg\\
% email address must be in roman text type, not monospace or sans serif
}

\maketitle

\begin{abstract}

This work is inspired by recent advances in hierarchical reinforcement learning (HRL)~\cite{recent-hrl-Barto2003,hrl-springer-Hengst2010}, and improvements in learning efficiency from heuristic-based subgoal selection, experience replay~\cite{experience-replay-Lin1993,HER-Andrychowicz2017}, and task-based curriculum learning~\cite{Bengio2009-curriculum-learning,Zaremba2014-curriculum-task-specific}. We propose a new method to integrate HRL, experience replay and effective subgoal selection through an implicit curriculum design based on human expertise to support sample-efficient learning and enhance interpretability of the agent’s behavior. Human expertise remains indispensable in many areas such as medicine~\cite{ai-medicine-Buch2018} and law~\cite{ai-law-Cath2018}, where interpretability, explainability and transparency are crucial in the decision making process, for ethical and legal reasons. Our method simplifies the complex task sets for achieving the overall objectives by decomposing them into subgoals at different levels of abstraction. Incorporating relevant subjective knowledge also significantly reduces the computational resources spent in exploration for RL, especially in high speed, changing, and complex environments where the transition dynamics cannot be effectively learned and modelled in a short time. Experimental results in two StarCraft II (SC2)~\cite{Vinyals2017} minigames demonstrate that our method can achieve better sample efficiency than flat and end-to-end RL methods, and provides an effective method for explaining the agent’s performance.

% \begin{description}
% Secondary publications and information retrieval purposes.
% You may use the \texttt{description} environment to structure your abstract;
% use the optional argument of the \verb+\item+ command to give the category of each item. 
% \end{description}
\end{abstract}

\section{\label{sec:introduction}Introduction}

Reinforcement learning (RL)~\cite{RLbook2018} enables agents to learn how to take actions, by interacting with an environment, to maximize a series of rewards received over time. In combination with advances in deep learning and computational resources, the Deep Reinforcement Learning (DRL) \cite{dqn-Mnih2013} formulation has led to dramatic results in acting from perception \cite{human-level-control-Mnih2015}, game playing \cite{go-Silver2016a}, and robotics \cite{robotics-OpenAI2018}. However, DRL usually requires extensive computations to achieve satisfactory performance. For example, in full-length StarCraft II (SC2) games, AlphaStar~\cite{alphaStar} achieves superhuman performance at the expense of huge computational resources\footnote{According to~\cite{alphaStar}, for each of their 12 agents, they conduct training on 32 TPUs for 44 days.}. Training flat DRL agents even on minigames (simplistic versions of the full-length SC2 games) requires 600 million samples~\cite{Vinyals2017} and 10 billion samples~\cite{deep-relational-Zambaldi2019} for each minigame, and repeated with 100 different sets of hyper-parameters, approximately equivalent to over 630 and 10,500 years of game playing time respectively. Even with such large number of training samples, DRL agents are not yet able to beat human experts at some minigames \cite{Vinyals2017,deep-relational-Zambaldi2019}.

We argue that learning a new task in general or SC2 minigames in particular is a two-stage process, viz., learning the fundamentals, and mastering the skills. For SC2 minigames, novice human players learn the minigame fundamentals reasonably quickly by decomposing the game into smaller, distinct and necessary steps. However, to achieve mastery over the minigame, humans take a long time, mainly to practice the precision of skills. RL agents, on the other hand, may take a long time to learn the fundamentals of the gameplay but achieve mastery (stage two) efficiently. This can be observed from the training progress curves in~\cite{Vinyals2017} which shows spikes followed plateaus of reward signals instead of steady and gradual increases.

We want to leverage human expertise to reduce the `warm-up' time required by the RL agents. The Hierarchical Reinforcement Learning (HRL) framework~\cite{HRL-subgoalsubpolicy-Bakker2004,HER-HRL-Evel2019} comprises a general layered architecture that supports different levels of abstractions corresponding to human expertise and agent's skills at the low-level manoeuvres. Intuitively, HRL provides a way for combining the best from human expertise and agent by organizing the inputs from humans at a high level (more abstract) and those from agents at a lower level (more precise). In this work, we extend the HRL framework to incorporate human expertise in subgoal selection. We demonstrate the effects of our methods in mastering SC2 minigames, and present preliminary results on sample efficiency and interpretability over the flat RL methods.

The rest of the paper is organized as follows. We briefly outline the background information in the next section. Next, we describe our proposed methodology. Further, we discuss the related works and present our experimental results. We then conclude the paper highlighting opportunities for future work.

\section{\label{sec:prelim}Preliminaries}

\textbf{Markov decision process and Reinforcement learning:} \\
% \begin{definition} Markov Decision Process (MDP) \\
% \label{def:mdp}
A Markov decision process (MDP) is a five-tuple $ \langle  \mathcal{S}, \mathcal{A}, \mathcal{T}, \mathcal{R}, \gamma \rangle$, where, $\mathcal{S}$ is the set of states the agent can be in; $\mathcal{A}$ is the set of possible actions available for the agent; $\mathcal{R}: \mathcal{S} \times\mathcal{A}\mapsto \mathbb{R}$ is the reward function, $\mathcal{T}: \mathcal{S} \times \mathcal{A} \mapsto \Delta \mathcal{S} $ is the transition function; and $\gamma \in [0, 1]$ is the discount factor that denotes the usefulness of the future rewards.
% \end{definition}
%
We consider the standard formalism of reinforcement learning where an agent continuously interacts with a fully observable environment, defined using an MDP. A deterministic policy is a mapping $\pi : \mathcal{S}\mapsto \mathcal{A}$ and we can describe a sequence of actions and reward signals from the environment. Every episode begins with an initial $s_0$. At each $t$, the agent takes an action $a_t = \pi_t (s_t)$, and gets a reward $r_t = \mathcal{R}(s_t, a_t)$. At the same time, $s_{t+1}$ is sampled from $\mathcal{T}(s_{t}, a_{t})$. Over time, the discounted cumulative reward, called \textit{return}, is calculated as: $R_t = \sum_{i=t}^{\infty} \gamma^{i-t} r_{t}$. The agent's task is to maximize the expected \textit{return} $\mathbb{E}_{s_0}[R_0|s_0]$. Furthermore, the Q-function (or action-value function) is defined as $Q^{\pi}(s_t, a_t)=\mathbb{E}[R_t|s_t, a_t]$. Assuming an optimal policy $\pi^*: Q^{\pi^*}(s, a) \geq Q^{\pi}(s, a) \: \forall s\in\mathcal{S}, a\in\mathcal{A}, \text{for any possible} \: \pi$. All optimal policies have the same Q-function called the \textit{optimal Q-function}, denoted $Q^*$, satisfying this Bellman equation: \[ Q^*(s, a) = \mathbb{E}_{s' \sim \mathcal{T}(s,a)}[\mathcal{R}(s,a)+ \gamma \max_{a'\in \mathcal{A}} Q^*(s', a')]. \]

\noindent\textbf{Q-function Approximators}
The above definitions enable one possible solution to MDPs: using a function approximator for $Q^*$. \textit{Deep-Q-Networks} (DQN) \cite{dqn-Mnih2013} and \textit{Deep Deterministic Policy Gradients} (DDPG) \cite{DDPG-Lillicrap2016}, are such approaches tackling model-free RL problems. Typically, a neural network $Q$ is trained to approximate $Q^*$. During training, experiences are generated via an \textit{exploration policy}, usually $\epsilon$-greedy policy with the current $Q$. The experience tuples $(s_t,a_t,r_t,s_{t+1})$ are stored in a \textit{replay buffer}. $Q$ is trained using gradient descent with respect to the loss $L \coloneqq \mathbb{E}[Q(s_t, a_t) - y_t]^2$, where $y_t = r_t + \gamma \max_{a'\in \mathcal{A}}Q(s_{t+1}, a')$ with experiences sampled from the \textit{replay buffer}.

An \textit{exploration policy} is a policy that describes how the agent interacts with the environment. For instance, a policy that picks actions randomly encourages \textit{exploration}. On the other hand, a \textit{greedy} policy with respect to $Q$, as in $\pi_Q(s) = \argmax_{a\in\mathcal{A}}Q(s,a)$, encourages exploitation. To balance these, a standard approach of $\epsilon$-greedy~\cite{RLbook2018} is adopted: with probability $\epsilon$ take a random action, and with probability $1-\epsilon$ take a \textit{greedy} action.

\noindent\textbf{Goal Space $\mathcal{G}$}
\citet{uvfa-Schaul2015} extended DQN to include a goal space $\mathcal{G}$. A (sub)goal can be described with specifically selected states, or via functions such as  $f: \mathcal{S} \mapsto [0, 1]$, either a state is a goal or not. Introducing $\mathcal{G}$ modifies the original reward function $\mathcal{R}$ slightly: $\forall g\in \mathcal{G}$, $\mathcal{R}_g: \mathcal{S}\times \mathcal{A} \mapsto \mathbb{R}, \; \mathcal{R}(s,a|g) \coloneqq \mathcal{R}_g(s,a)$. At the beginning of each episode, in addition to $s_0$, the initialization includes a fixed $g$ to create a tuple $(s_0, g)$. Other modifications naturally follow: $\pi:\mathcal{S}\times\mathcal{G}\mapsto \mathcal{A}$, and $Q^{\pi}(s_t, a_t, g) = \mathbb{E}[\mathcal{R}_t|s_t, a_t, g]$.

\noindent\textbf{Experience Replay}
\citet{experience-replay-Lin1993} proposed the idea of using `experiences buffers' to help machines learn. Formally, a single time step experience is defined as a tuple $(s_t,a_t,r_t,s_{t+1})$ and more generally an experience can be constructed by concatenating multiple consecutive experience tuples.

\noindent\textbf{Curriculum Learning}
Methods in this framework typically explicitly or implicitly design a series of tasks or goals (with gradually increased difficulties) for the agent to follow and learn, i.e., the curriculum~\citep{Bengio2009-curriculum-learning,weng2020curriculum-blog}.

\noindent\textbf{StarCraft II}
SC2 is a real-time-strategy (RTS) game, where players command their units to compete against each other. In an SC2 full-length game, typically players start out by commanding units to collect resources (minerals and gas) to build up their economy and army at the same time. When they have amassed a sufficiently large army, they command these units to attack their opponents' base in order to win. SC2 is currently a very promising simulation environment for RL, due to its high flexibility and complexity and wide-ranging applicability in the fields of game theory, planning and decision making, operations optimization, etc. SC2 minigames, as opposed to full-length games described above, are built-in episodic tutorials where novice players can learn and practice their skills in a controlled and less complex environment. Some relevant skills include collecting resources, building certain army units, etc.

\section{\label{sec:proposed}Proposed Methodology}
We propose a novel method of integrating the advantages of human expertise and RL agents to facilitate fundamentals learning and skills mastery of a learning task. Our method adopts the principle of \textit{Curriculum Learning}~\citep{Bengio2009-curriculum-learning} and follows a task-oriented approach~\citep{Zaremba2014-curriculum-task-specific}. The key idea is to leverage human expertise to simplify the complex learning procedure, by decomposing it into hierarchical subgoals as the curriculum for the agent. More specifically, we factorize the learning task into several successive subtasks indispensable for the agent to complete the entire complex learning procedure. The customized reward function in each subtask implicitly captures the corresponding subgoal. Importantly, these successive subgoals are determined so that they are gradually more difficult to improve learning efficiency~\cite{Bengio2009-curriculum-learning,Justesen2018-illuminating-increasing-difficulty}. With defined subgoals, we use the \textit{Experience Replay} technique to construct the experiences to further improve the empirical sample efficiency~\cite{HER-Andrychowicz2017,HRL-subgoalsubpolicy-Bakker2004,HER-HRL-Evel2019}. Furthermore, adopting clearly defined subtasks and subgoals enhances the interpretability of the agent's learning progress. In implementation, we customize SC2 minigames to embed human expertise on subgoal information and the criteria to identify and select subgoals during learning. Therefore, the agent learns the subpolicies and combines them in a hierarchical way. By following a well-defined decomposition of the original minigame into subtasks, we can choose the desired state of a previous subtask to be the starting conditions of the next subtask, thus completing the connection between subtasks.

\subsection{Hierarchy: Subgoals and Subtasks}
Our proposed hierarchy is composed of subgoals, which collectively divide the problem into simpler subtasks that can be solved easily and efficiently. Each subgoal is implicitly captured as the desired state in its corresponding subtask, and we refer to the agent's skills to reach a subgoal its corresponding subpolicy. The rationale behind this is as follows. First, the advantages of human expertise and the agents are complementary to each other in terms of learning and mastering the task. Human players are good at seeing the big picture and thus identifying the essential and distinct steps/skills very quickly. On the other hand, agents are proficient in honing learned skills and maneuvers to a high degree of precision. Second, a hierarchy helps reduce the complexity of search space via divide-and-conquer. Lastly, this method enhances the interpretability of the subgoals (and subpolicies).

\begin{figure}[!ht]
\includegraphics[width=1\columnwidth]{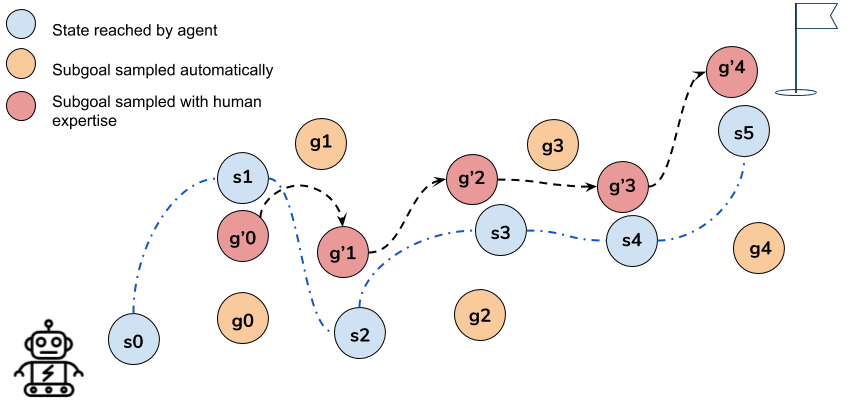}
\caption{\label{fig:nav-agent} Navigation Agent}
\end{figure}

Figure~\ref{fig:nav-agent} illustrates the concept of subgoals and subpolicies with a simple navigation agent. The agent is learning to navigate to the flag post from the initial state $s_0$. One possible sequence of the states is $s_1,\ldots,s_5$. Therefore, the entire trajectory can be decomposed into subgoals; for instance, \citet{HER-HRL-Evel2019} used heuristic-based subgoal selection criteria (in Figure~\ref{fig:nav-agent} these selected subgoals, $g_0,\ldots,g_4$, are denoted by orange circles). On the other hand, the sequence of red nodes denote subgoals of our method. We highlight that this sequence would constitute a better guided and more efficient exploration path. In addition this sequence is better aligned with the game where some states are the prerequisites for other states~(illustrated as the black dashed arrows).

\begin{figure}[!ht]
\minipage{0.23\textwidth}
    \includegraphics[width=\linewidth]{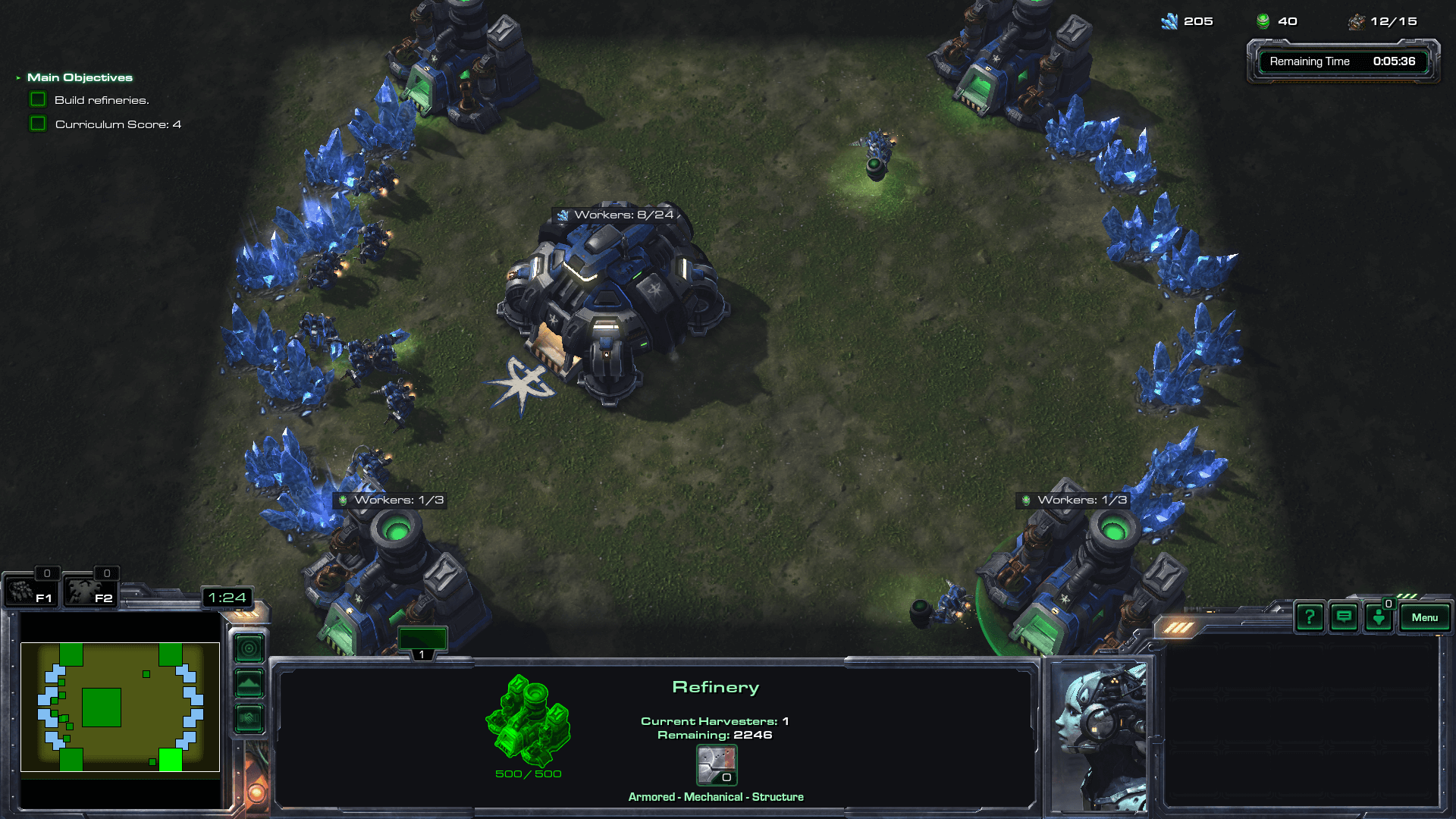}

\endminipage\hfill
\minipage{0.23\textwidth}
  \includegraphics[width=\linewidth]{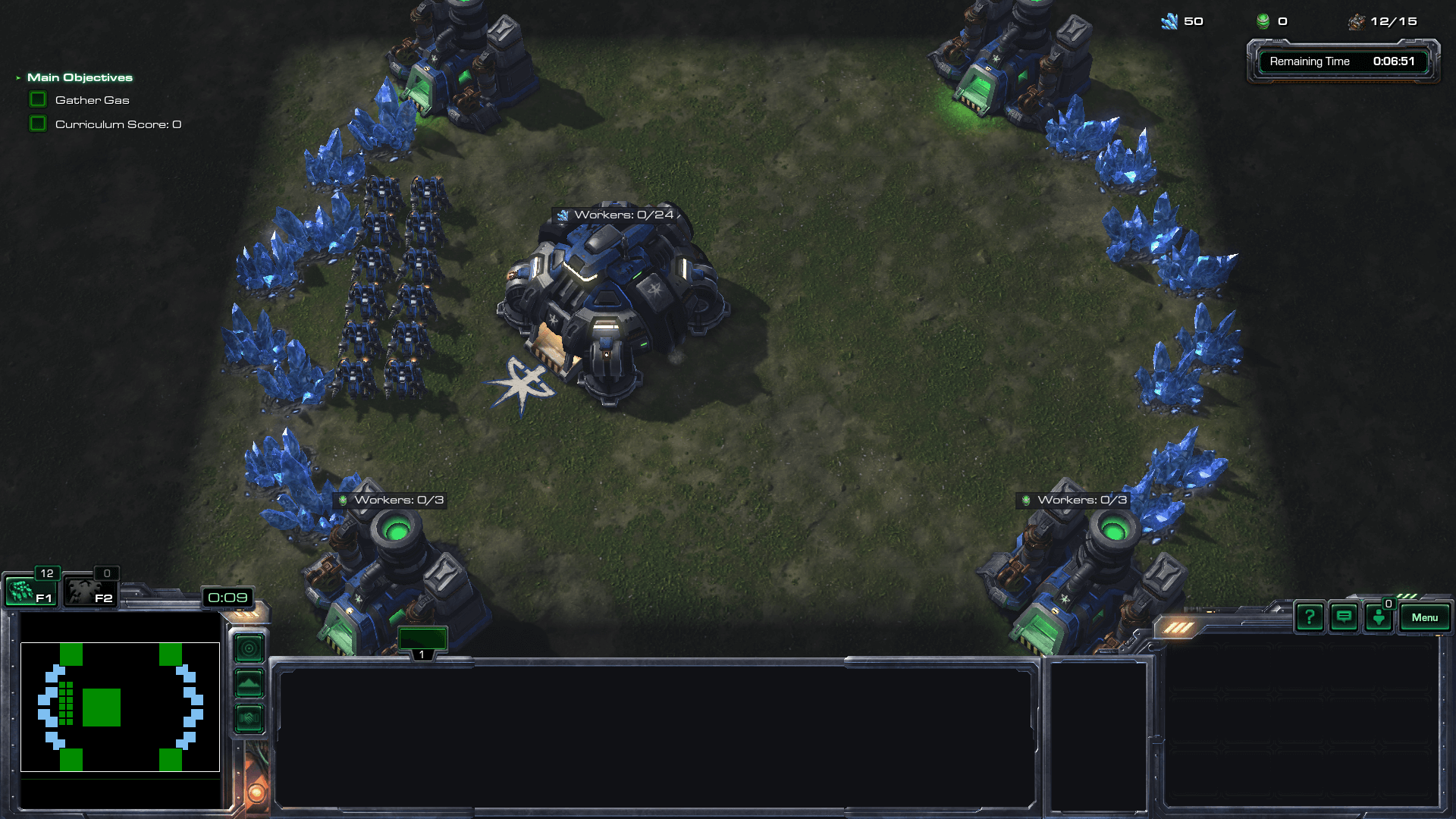}
\endminipage\hfill
\minipage{0.23\textwidth}%
  \includegraphics[width=\linewidth]{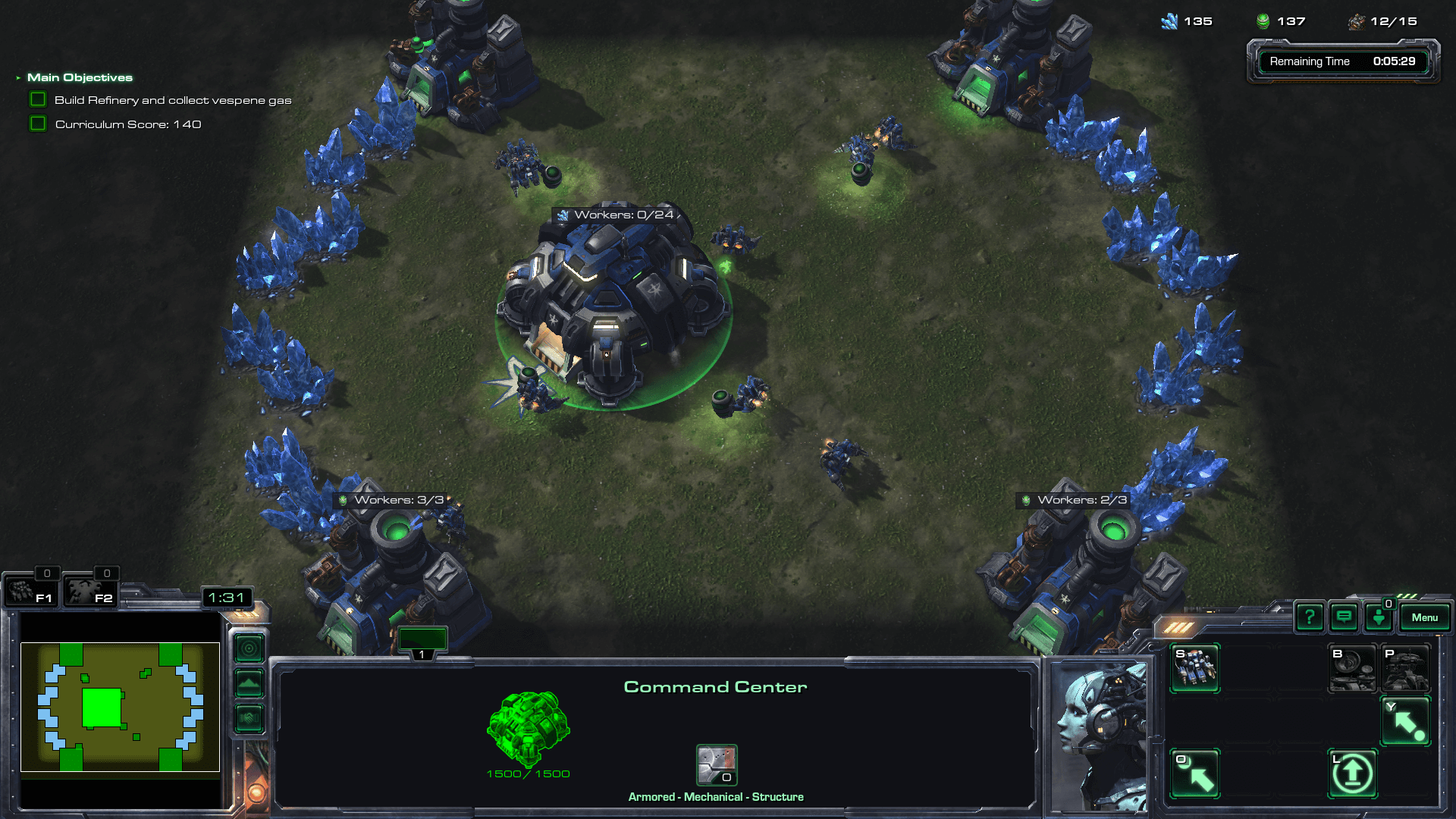}
\endminipage\hfill
\minipage{0.23\textwidth}%
  \includegraphics[width=\linewidth]{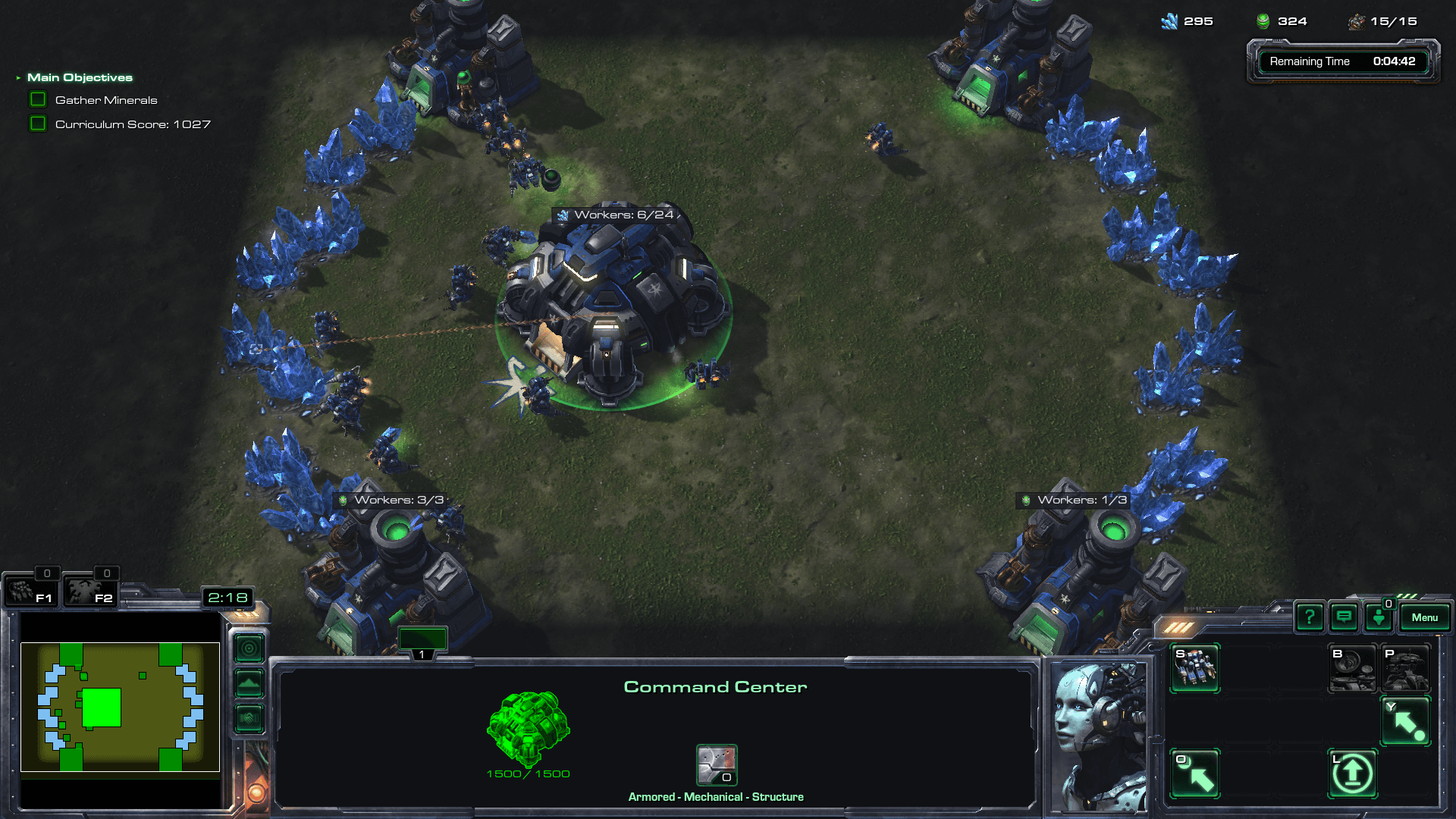}
\endminipage
\caption{\label{fig:sc2-minigames-collectresources}Collect Minerals and Gas. From left to right, top to bottom:(1)-(4):~(1) to build refineries; (2) to collect gas with built refineries; (3) both tasks in (1) and (2); (4) all three tasks in (1), (2), (3) and collect minerals.}
\end{figure}

\begin{figure}[!ht]
\minipage{0.23\textwidth}
  \includegraphics[width=\linewidth]{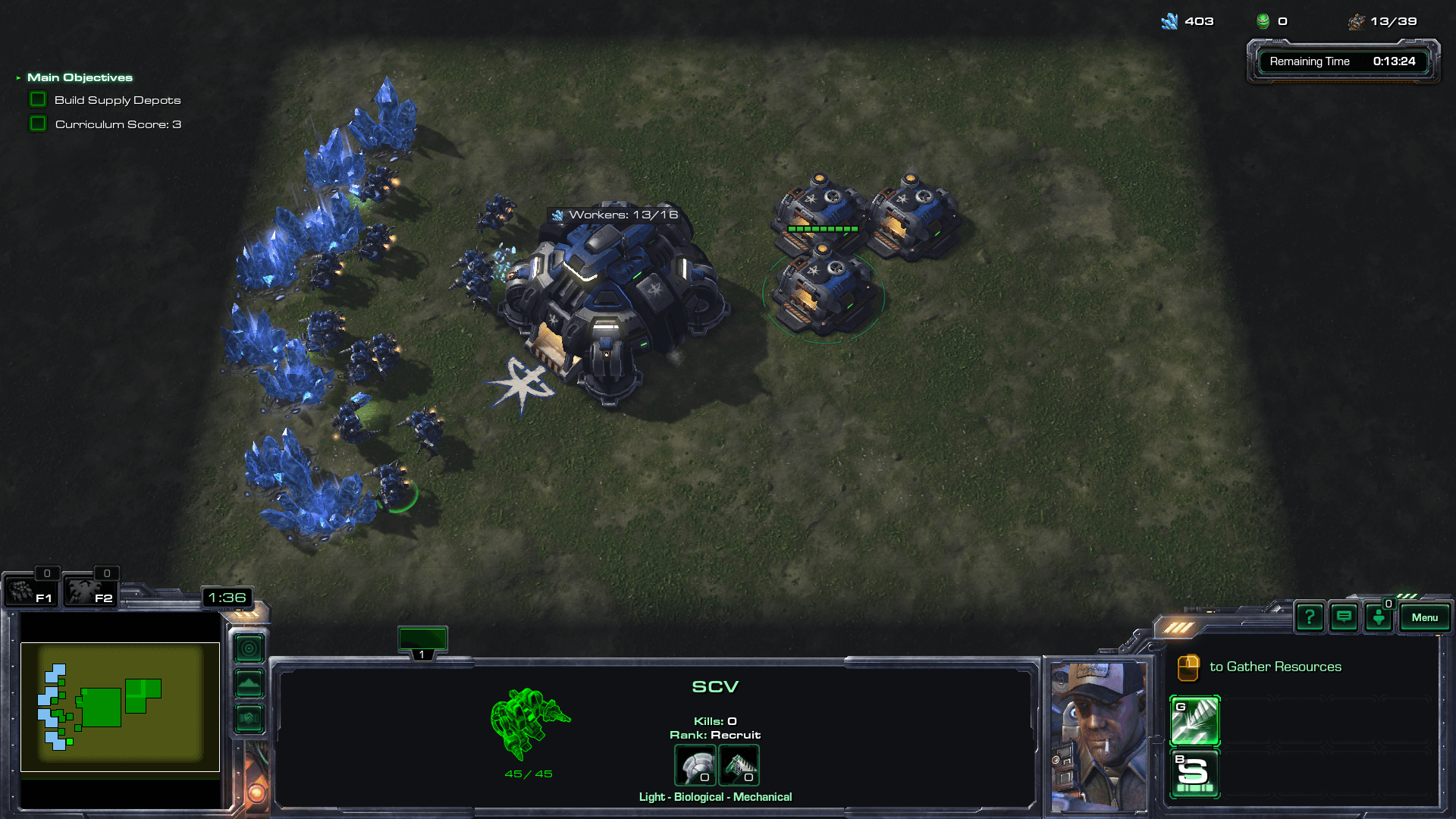}
\endminipage\hfill
\minipage{0.23\textwidth}
  \includegraphics[width=\linewidth]{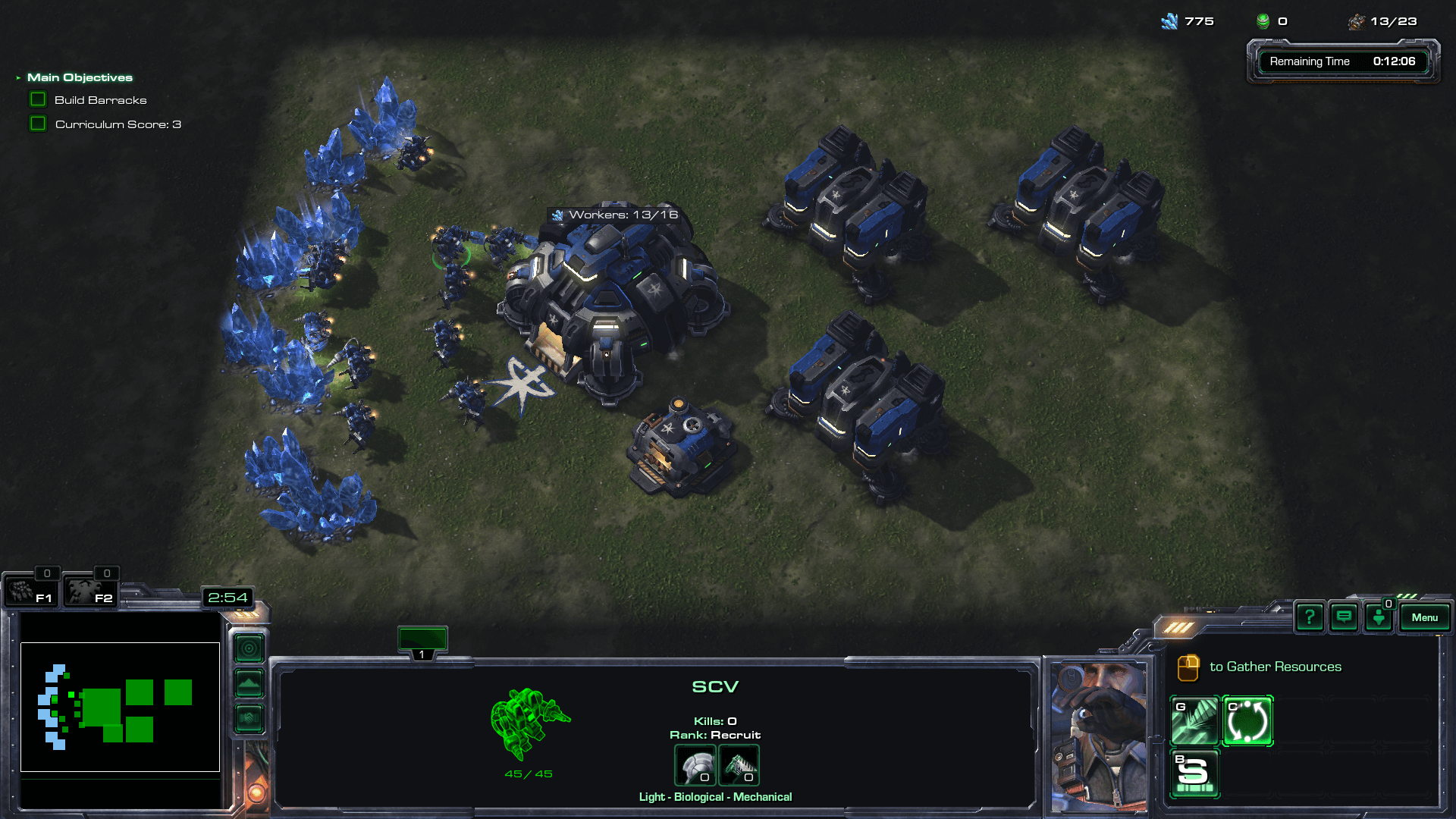}
\endminipage\hfill
\minipage{0.23\textwidth}%
  \includegraphics[width=\linewidth]{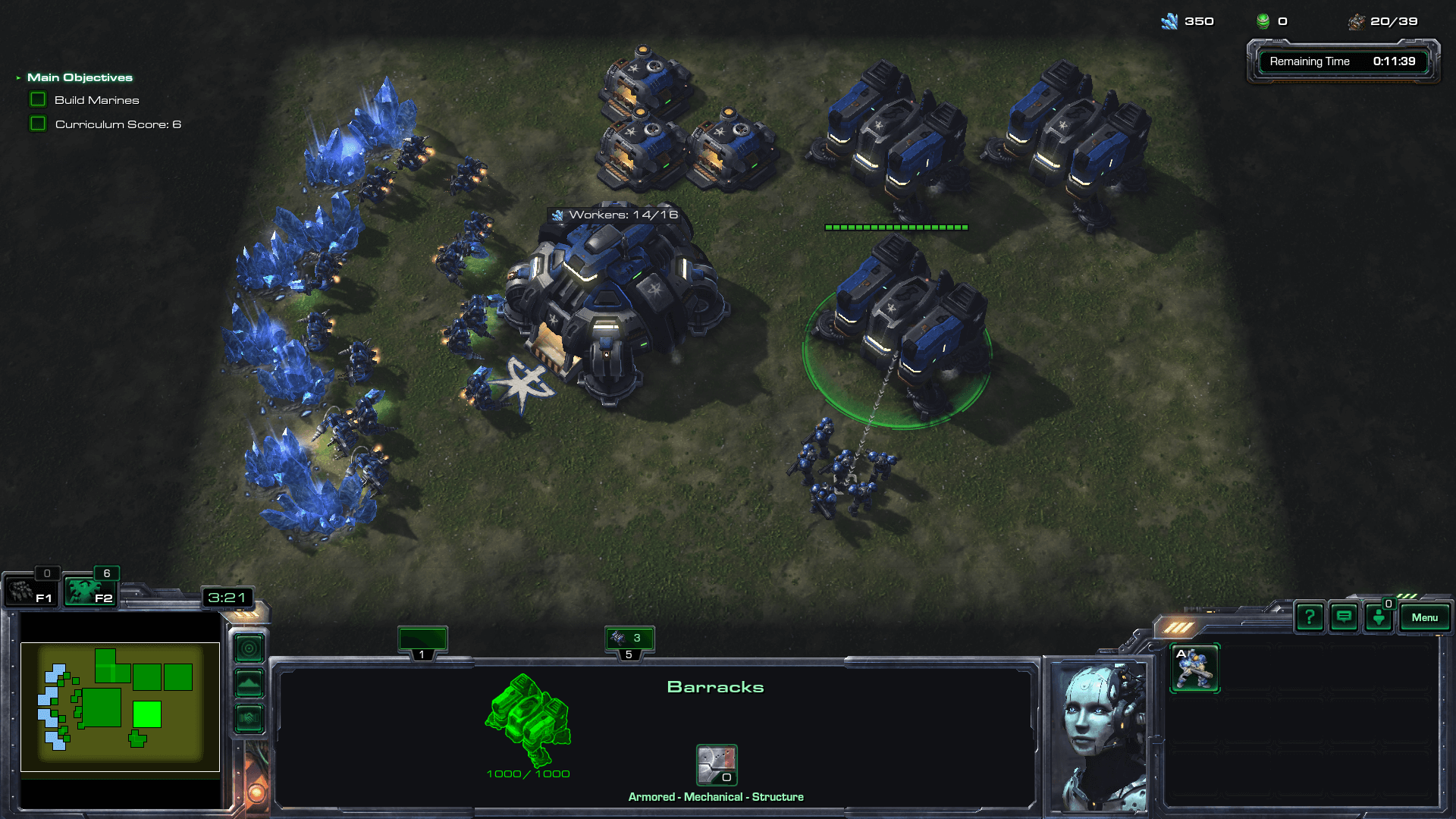}
\endminipage\hfill
\minipage{0.23\textwidth}%
  \includegraphics[width=\linewidth]{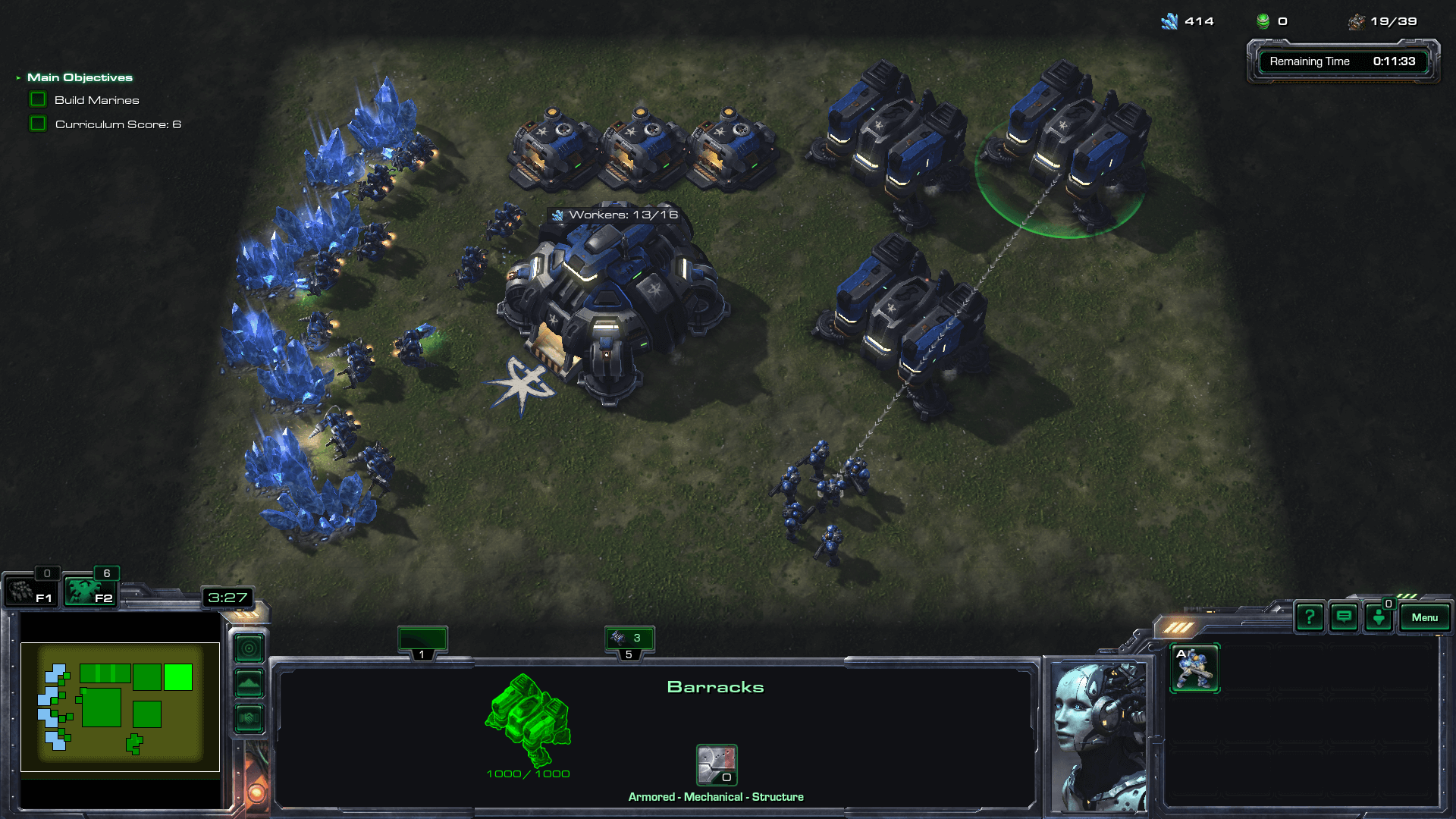}
\endminipage
\caption{\label{fig:sc2-minigames-buildmarines} Build Marines. From left to right, top to bottom:(1)-(4):~(1) to build supply depots; (2) to build barracks; (3) to build marines with (1) and (2) already built; (4) all three tasks in (1), (2), (3). }
\end{figure}

\subsection{Subgoals Selection and Experience Replays}
\textit{Subgoal Design and Selection}. We use the similar method for constructing experiences with a goal space as previous works~\cite{HER-Andrychowicz2017,HER-HRL-Evel2019}. However, our method introduces human expertise in constructing the hierarchy and subgoals selection. In \cite{HER-Andrychowicz2017}, the \textit{hindsight experience replay} buffer is constructed via random sampling from the goal space and concatenating the sampled goals to an \textit{already executed} sequence $\{s_1,\ldots,s_T\}$, hence the name hindsight. The subgoals are initialized with heuristic-based selection and updated according to \textit{hindsight actions}. For example, in Figure~\ref{fig:nav-agent}, given a predetermined subgoal $g_0$, the agent might not successfully reach it, and instead ends up in $s_1$. In this case, the subgoal set in hindsight is $s_1$ (updated from $g_0$).

Our method distinguishes in that the (sub)goals selection strategy is designed with human expertise, to give a fixed but suitable decomposition of the learning task. Furthermore, we exploit the underlying sequential relationship among the subgoals as in the game some states are the prerequisites for others. Hence, certain actions are required to be performed in order. Furthermore, another reason for introducing human expertise rather than using end-to-end learning alone is that compared with the environments investigated in previous HRL works, SC2 encompasses a significantly larger state-action space that prohibits a sample-efficient end-to-end learning strategy. As a result, our method enjoys an added advantage of interpretability of the selected subgoals.

\textit{Subtasks Implementations}. We leverage the customizability of SC2 minigames to carefully design subtasks to enable training of the corresponding subpolicies, as suggested in~\cite{recent-hrl-Barto2003}. We illustrate with the \textit{Collect Minerals and Gas} (CMAG) minigame, as shown and described in Figure~\ref{fig:sc2-minigames-collectresources}. There are several distinct and sequential actions the player has to perform to score well: 1. commanding the Space Construction Vehicles (SCVs) - basic units of the game, to collect minerals; 2. having collected sufficient minerals, selecting SCVs to build the gas refinery (a prerequisite building for collecting vespene gas) on specific locations with existing gas wells; 3. commanding the SCVs to collect vespene gas from the constructed gas refinery; 4. producing additional SCVs (at a fixed cost) to optimize the mining efficiency. And there is a fixed time duration of 900 seconds. The challenge of CMAG is that all these actions/subpolicies should be performed in an optimized sequence for best performance. The optimality depends on the order, timing, and the number of repetitions of these actions. For instance, it is important not to under/over-produce SCVs at a mineral site for optimal efficiency. Hence, we implemented the following subtasks: \textit{BuildRefinery}, \textit{CollectGasWithRefineries} and \textit{BuildRefinieryAndCollectGas}. In the first two subtasks, the agent learns the specific subpolicies to build refineries and to collect gas~(from built refineries), respectively, while in the last subtask the agent learns to combine them. Based on the same idea, the complete decomposition for CMAG is given by [CMAG, \textit{BuildRefinery}, \textit{CollectGasWithRefineries}, \textit{BuildRefineryAndCollectGas}, CMAG] where the first CMAG trains the agent to collect minerals, and the last CMAG trains it to combine all subpolicies and also `re-introduces' the reward signal for collecting minerals to avoid forgetting~\cite{Zaremba2014-curriculum-task-specific}. Similarly, for the BuildMarines (BM) minigame, shown in Figure~\ref{fig:sc2-minigames-buildmarines}, the sequential steps/actions are: 1. commanding the SCVs to collect minerals; 2. having collected sufficient minerals, selecting SCVs to build a supply depot (a prerequisite building for barracks and to increase the supplies limit); 3. having both sufficient minerals and a supply depot, selecting SCVs to build barracks; 4. having minerals, a supply depot and barracks and with current unit count less than the supplies limit, selecting the barracks to train marines. The fixed time duration for BM is 450 seconds. Therefore, we implemented the corresponding subtasks: \textit{BuildSupplyDepots}, \textit{BuildBarracks}, \textit{BuildMarinesWithBarracks} and the complete decomposition for BM is [\textit{BuildSupplyDepots}, \textit{BuildBarracks}, \textit{BuildMarinesWithBarracks}, BM]. Note we do \textit{not} set BM as a first subtask as for CMAG because CMAG contains both reward signals for minerals and gas, so it is an adequate simple task for the agent to learn to collect minerals. However, BM has only the reward signals for training marines, thus too difficult as the first subtask.

\textit{Construct Experience Replay for Each Subtask}. With the designed subtasks represented by our customized minigames, constructing experience replays is straightforward. For a subtask, a predetermined subgoal $g_i$ is implicitly captured in its customized minigame (e.g., to build barracks, to manufacture SCVs, etc.) using a corresponding reward signal, so that the agent learns to reach $g_i$. For the immediate subsequent subtask, we set its initial conditions to be the completed subgoal $g_i$. So, the agent learns to continue on the basis of a completed $g_i$. It is an implicit process because, when learning to reach subgoal $g_{i+1}$, the agent does \textit{not} see or interact directly with the reward signal corresponding to $g_i$. For example, between two ordered subtasks \textit{CollectMinerals} and \textit{BuildRefinery}, the agent learns to collect minerals first and starts with some collected minerals in the latter with the sole objective of learning to build refineries.

\textit{Off-policy learning and }PPO. Off-policy learning is a learning paradigm where the exploration and learning are decoupled and take place separately. Exploration is mainly used by the agent to collect experiences or `data points' for its policy function or model. Learning is then conducted on these collected experiences, and Proximal Policy Optimization (PPO)~\citep{PPO-Schulman2017} is one such method. Its details are not the focus of this work and omitted here.

\textit{Algorithm}. We describe the HRL algorithm with human expertise in subgoal selection here. The pseudo-code is given in Algorithm~\ref{alg:alg1}. For a learning task, a sequence of subtasks is designed with human expertise to implicitly define the subgoals and we refer to our customized SC2 minigames as subtasks $\Gamma_i, 0 \leq i < m$ for the learning task. We pre-define reward thresholds $thresholds \in \mathbb{R}^m$, for all subtasks. As the agent's running average reward is higher than a threshold, this agent is considered to have learnt the corresponding subtask well and will move to the subsequent subtask. We use learner $\mathcal{L}$ to denote the agent and to describe how it makes decisions and takes actions. It can be represented by a deep neural network, and parametrized by $\vec{w}_{\mathcal{L}}$. In addition, we define a sample count $c$ and sample limit $n$. Sample count $c$ refers to the number of samples the agent has used for learning a subtask. Sample limit $n$ refers to the total number of samples allowed for the agent for the entire learning task, i.e., for all subtasks combined. $c$ and $n$ together are used to demonstrate empirical sample efficiency. 

With these definitions and initializations, the algorithm takes the defined sequence of subtasks $\Gamma$ with corresponding $thresholds$ and initiates learning on these subtasks in the same sequence. During the process, a running average of the agent's past achieved rewards is kept for each subtask, represented by the API call \texttt{test()}. For each subtask $\Gamma_i$, either the agent completely exhausts its assigned sample limit $\lfloor \frac{n}{m} \rfloor$ or it successfully reaches the $thresholds_i$. If the running average of past rewards $\geq thresholds_i$, the agent completes learning on $\Gamma_i$ and starts with $\Gamma_{i+1}$; the process continues until all subtasks are learned. We follow the \textit{exploration policy} in preliminaries and adopt an $\epsilon$-greedy policy, represented by \texttt{explore()} in Algorithm~\ref{alg:alg1}.

\begin{algorithm}
\caption{HRL with Human Expertise in Subgoal Selection}
\label{alg:alg1}
\small
\begin{algorithmic} 
\Require subtasks $\Gamma_i, 0 \leq i < m$
\Require reward thresholds $thresholds \in \mathbb{R}^m$ \Require learner $\mathcal{L}$, parametrized by $\vec{w}_{\mathcal{L}}$
\Require sample count $c$, sample limit $n$.
    \For{$ 0 \leq i < m $}
        \State $c \gets 0$
        \While{$ c <= \lfloor \frac{n}{m} \rfloor $}
            \State $ experiences \gets $ \texttt{explore}$(\mathcal{L}, \Gamma_i)$
            
            \State $c \gets c + |experiences|$

            \State $\vec{w}_{\mathcal{L}}' \gets $\texttt{PPO}$(\vec{w}_{\mathcal{L}}, experiences)$  \Comment{off-policy}

            \If {\texttt{test($\vec{w}_{\mathcal{L}}'$) $\geq thresholds_i$ }}
            \State Break \Comment{Go to next subtask}
            \EndIf

        \EndWhile
    
    \EndFor
    \vspace{2mm}
\end{algorithmic}
\end{algorithm}

\section{\label{sec:related}Related Work}

\textbf{Experience Replay} RL has achieved impressive developments in robotics \cite{roboticRL}, strategic games such as Go \cite{alphaGo}, real-time strategy games \cite{deep-relational-Zambaldi2019,alphaStar} etc. Researchers have attempted in various ways to address the challenge of goal-learning, reward shaping to get the `agent' to learn to master the task, and yet not overfit to the particular instances of the goals or reward signals. \textit{Experience Replay} \cite{experience-replay-Lin1993} is a technique to store and re-use past records of executions (along with the signals from the environment) to train the `agent', achieving efficient sample usage. \citet{dqn-Mnih2013} employed this technique together with Deep-Q-Learning to produce state-of-the-art results in Atari, and subsequently \citet{human-level-control-Mnih2015} confirmed the effectiveness of such approach under the stipulation that the `agent' only sees what human players would see, i.e., the pixels from the screen and some scoring indices. %Following that, \citet{HER-Andrychowicz2017} extended \textit{Experience Replay} into \textit{Hindsight Experience Replay} (HER), as a way for the `agent' to learn from reached goals, and more importantly, missed ones, thereby improving sample efficiency and making training possible in complex environments with only sparse and binary reward signals.

\textbf{Curriculum Learning} \citet{Bengio2009-curriculum-learning} hypothesized and empirically showed that introducing gradually more difficult examples speeds up the online learning, using a manually designed task-specific curriculum. \citet{Zaremba2014-curriculum-task-specific} experimentally showed that it is important to mix in easy tasks to avoid forgetting. \citet{Justesen2018-illuminating-increasing-difficulty} demonstrated that training an RL agent over a simple curriculum with gradually increasing difficulty can effectively prevent overfitting and lead to better generalization.

\textbf{Hierarchical Reinforcement Learning (HRL)} HRL and its related concepts such as \textit{options}~\cite{options-Sutton1999} \textit{macro-actions}~\cite{macro-actions-Hauskrecht2013}, or \textit{tasks}~\cite{CSRL-zhuoru2017-medcomp} were introduced to decompose the problem, usually a Markov decision process~(MDP), into smaller sub-parts to be efficiently solved. We refer the readers to~\cite{recent-hrl-Barto2003,hrl-springer-Hengst2010} for more comprehensive treatments. We describe two tracks of related works most relevant to our problem. \citet{HRL-subgoalsubpolicy-Bakker2004} proposed a two-level hierarchy, using \textit{subgoal} and \textit{subpolicy} to describe the learning taking place at the lower level of the hierarchy. \citet{HER-HRL-Evel2019} further articulated these ideas, and explicitly combined them with \textit{Hindsight Experience Replay}~\cite{HER-Andrychowicz2017} for better sample efficiency and performance. Another similarly inspired approach called \textit{context sensitive reinforcement learning}~(CSRL) introduced by~\citet{CSRL-zhuoru2017-medcomp} employed the hierarchical structure to enable effective re-use of learnt knowledge of similar (sub)tasks in a probabilistic way. In CSRL, instead of \textit{Experience Replay}, efficient simulations over constructed states are used in learning, able to learn both the tasks, and the environment (the transition and reward functions). CSRL scales well with state space, and is relatively easily parallelizable.

\textbf{StarCraft II} In addition to~\cite{deep-relational-Zambaldi2019}, several works addressed some of the challenges presented by SC2. In a real-time strategy~(RTS) game such as SC2, the hierarchical architecture is an intuitive solution concept, for its efficient representation and interpretability. Similar but different hierarchies were employed in two other works, where~\citet{modular-sc2-lee2018} designed the hierarchy with semantic meaning and from a operational perspective while~\citet{SC2Fulllength-Pang2018} forewent explicit semantic meanings for higher flexibility. Both provided promising empirical results on the full-length games against built-in AIs. Instead of full-length SC2 games, our investigation targets the minigames and we propose a way to integrate human expertise, the \textit{Curriculum Learning} paradigm and the \textit{Experience Replay} technique into the learning process.

Different from related works, our work adopts a principle-driven HRL approach with human expertise in the subgoal selection and thus an implicit formulation of a curriculum for the agent, on SC2 minigames in order to achieve empirical sample efficiency and to enhance interpretability.

\section{\label{sec:experiments}Experiments}
In the experiments, we specifically focus on two minigames, viz., BM and CMAG to investigate the effectiveness of our method. We choose these two because, the discrepancies in the performance between trained RL agents and human experts are the most significant as reported in~\cite{Vinyals2017}, suggesting these two are the most challenging for non-hierarchical end-to-end learning approaches. For both CMAG and BM, we have implemented our customized SC2 minigames~(subtasks) as described in the proposed methodology section, and we pair them with pre-defined reward thresholds. In our experiments, the decompositions for BM and CMAG are [\textit{BuildSupplyDepots}, \textit{BuildBarracks}, \textit{BuildMarinesWithBarracks}, BM], and [CMAG, \textit{BuildRefinery}, \textit{CollectGasWithRefineries}, \textit{BuildRefineryAndCollectGas}, CMAG], respectively.

\subsection{Experimental Setup}
\begin{itemize}
    \item \textit{Model Architecture and Hyperparameters}. We follow the model architecture of \textit{Fully Convolutional agent} in \cite{Vinyals2017} by utilizing an open-source implementation by \citet{reaver}. We use the hyperparameters listed in Table~\ref{table:hyperparameters}.
    \item \textit{Training \& Testing}.  In order to evaluate the empirical sample efficiency of our method, we restrict the total number of training samples to be 10 million. Note this is still significantly fewer than 600 million in \cite{Vinyals2017} or 10 billion in \cite{deep-relational-Zambaldi2019}. Furthermore, we adopt their practice of training multiple agents to report the best results attained. After training, on the trained model, average and maximum scores over 30 independent episodes are reported.
    \item \textit{Computing Resource}. CPU: Intel(R) Core(TM) i9-10920X CPU @ 3.50GHz, RAM:64 GB, GPU: GeForce RTX 2080 SUPER 8GB. The training time for a single model initialization: approximately 1.66 hours for CMAG and 1.5 hours for BM.
\end{itemize}

\begin{table}[!ht]
\caption{Hyperparameters}
\resizebox{\linewidth}{!}{%
\begin{tabular}{|c|c|c|}
    \hline
        & BM  &  CMAG \\\hline
    Learning rate & 0.0007 & 0.0007 \\ \hline
    Batch size &  32 & 32 \\ \hline
    Trajectory length & 40 & 40  \\ \hline
    Off-policy learning algorithm & PPO & PPO  \\ \hline
    Reward thresholds & [7,7,7,2] & [300,5,5,5,500] \\ \hline
\end{tabular}
\label{table:hyperparameters}
}
\end{table}

\begin{table}[!ht] 
\caption{Average Rewards Achieved}
\resizebox{\linewidth}{!}{%
\begin{tabular}{|c|c|c|c|c|}
    \hline
    Minigame & SC2LE & DRL  & Ours & Human Expert  \\ \hline
    CMAG & 3,978  &	5,055 &  478.5(527) & 7,566    \\ \hline
    BM   & 3        &	123   &  6.7(6.24) & 133    \\   \hline
\end{tabular}
\label{tbl:average_reward}
}
\end{table}

\begin{table}[!ht]
\caption{Maximum Rewards Achieved}
\resizebox{\linewidth}{!}{%
\begin{tabular}{|c|c|c|c|c|}
    \hline
    Minigame & SC2LE & DRL  & Ours & Human Expert \\ \hline
    CMAG & 4,130  &	 unreported   &  1825  & 7,566   \\ \hline
    BM   & 42     &  unreported   &  22 & 133    \\   \hline
\end{tabular}
\label{tbl:max_reward}
}
\end{table}

\begin{table}[!ht]
\caption{Training Samples Required}
\resizebox{\linewidth}{!}{%
\begin{tabular}{|c|c|c|c|c|}
    \hline
    Minigame & SC2LE & DRL  & Ours & Human Expert \\ \hline
    CMAG & 6e8    &	1e10  &  1e7  & N.A   \\ \hline
    BM &   6e8    &	1e10  &  3.4e6  & N.A   \\ \hline
\end{tabular}
\label{tbl:samples_required}
}
\end{table}

\begin{figure}[ht]
\minipage{0.49\textwidth}
    \includegraphics[width=\linewidth]{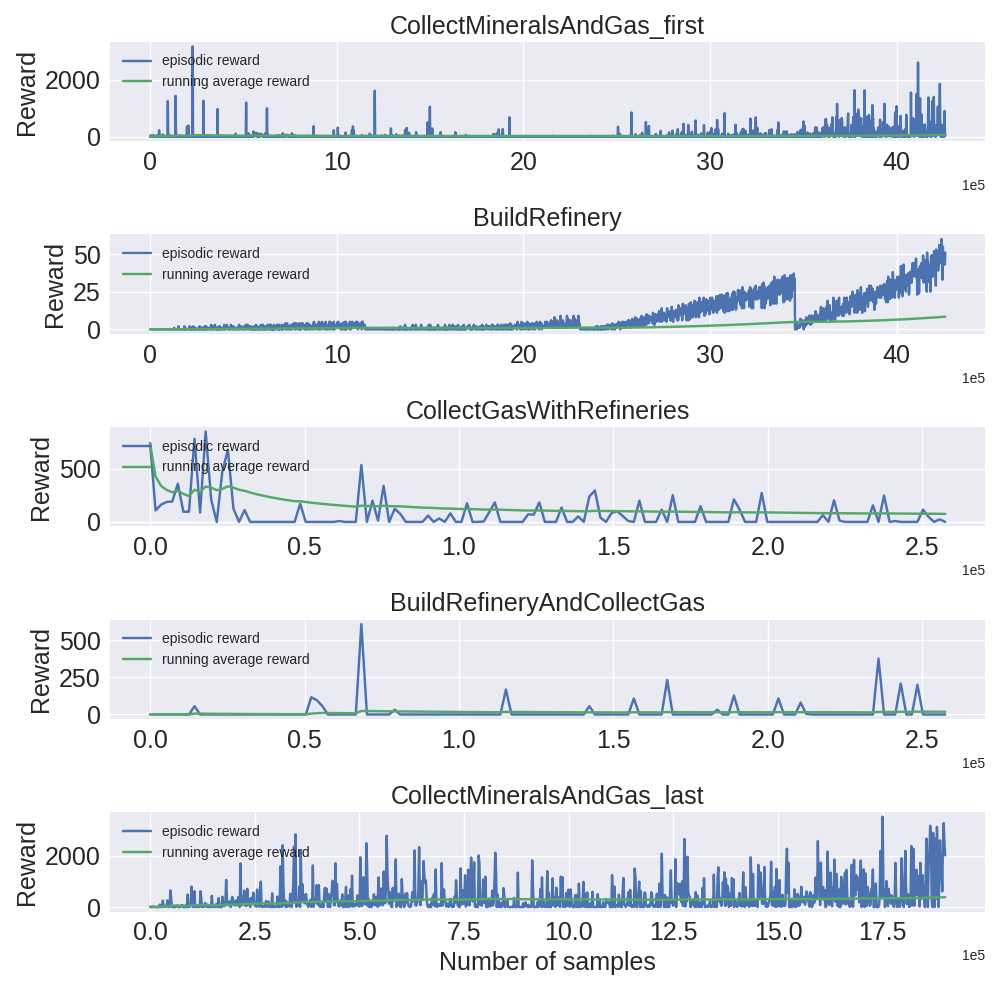}
\endminipage\hfill
\caption{\label{fig:CR-curve} Collect Minerals And Gas learning curve.}
\end{figure}

\begin{figure}[ht]
\minipage{0.49\textwidth}
    \includegraphics[width=\linewidth]{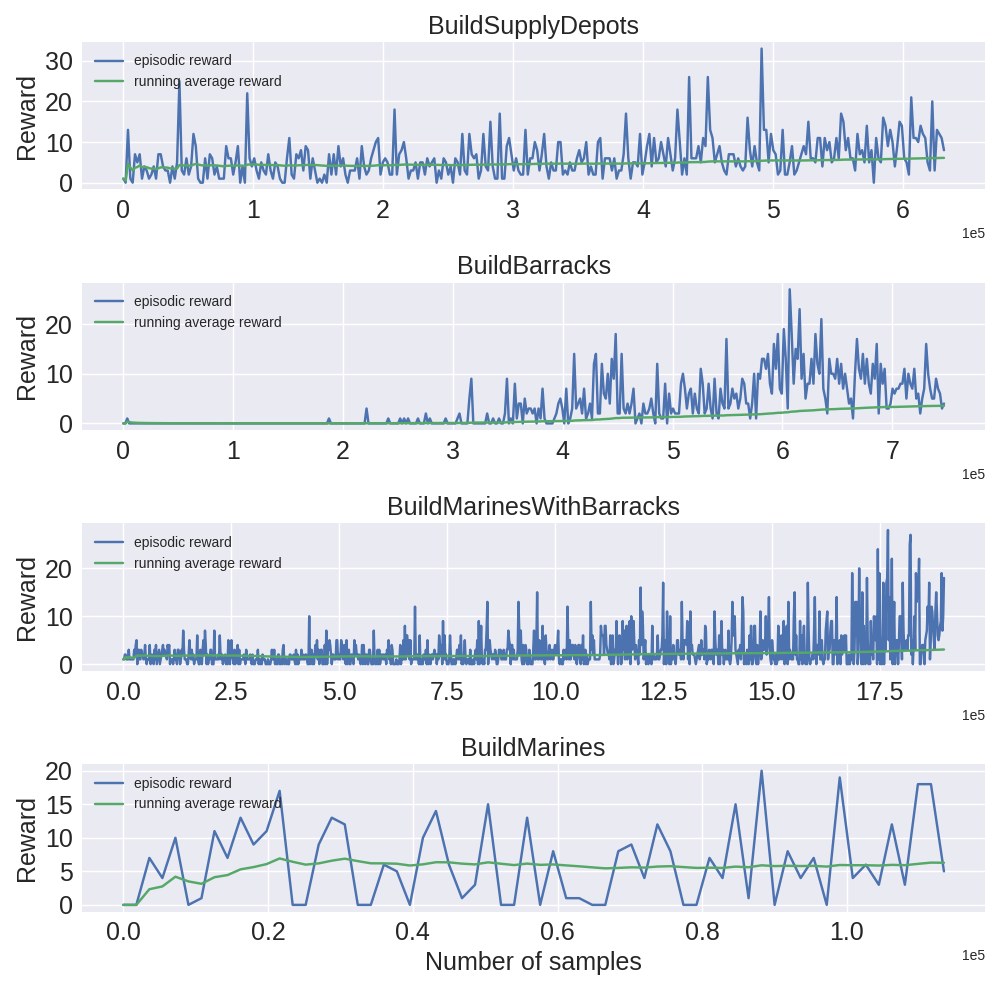}

\endminipage\hfill
\caption{\label{fig:BM-curve-best} Build Marines learning curve (best agent).}
\end{figure}

\begin{figure}[!ht]
\minipage{0.49\textwidth}
  \includegraphics[width=\linewidth]{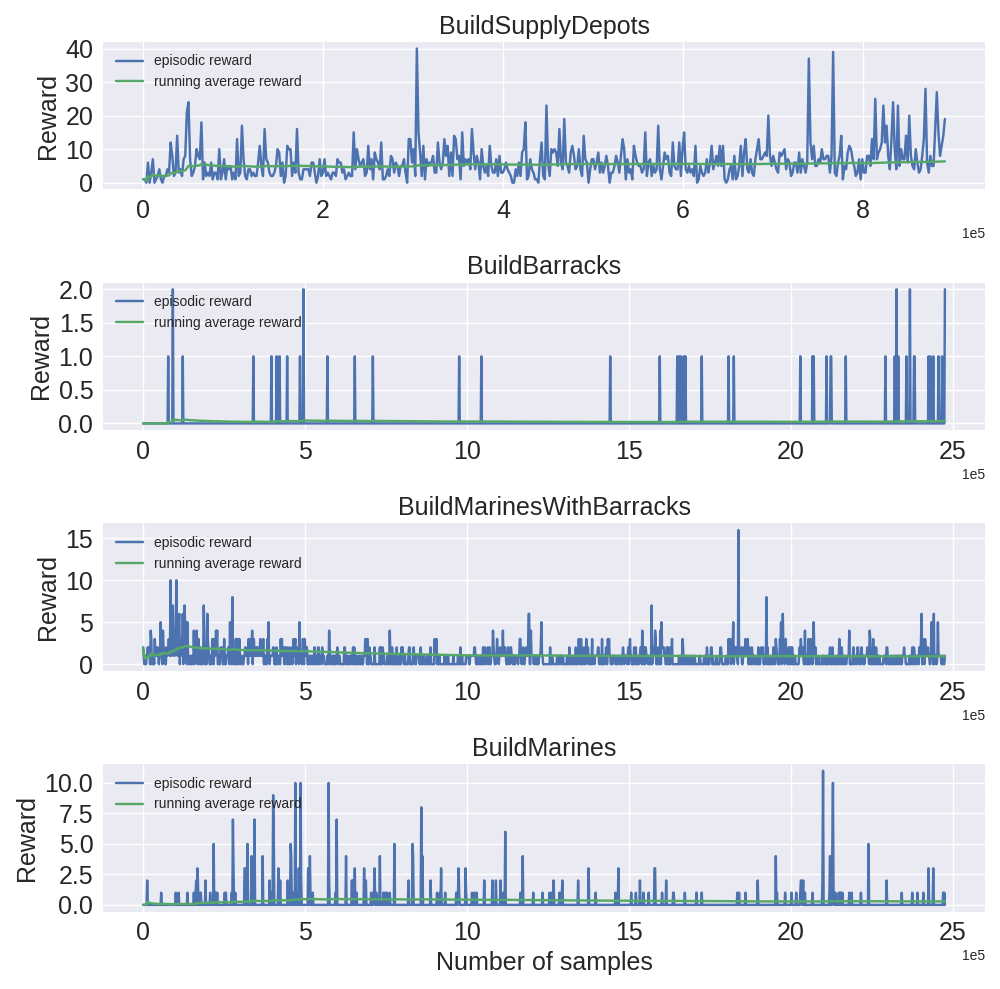}
\endminipage\hfill
\caption{\label{fig:BM-curve-worst} Build Marines learning curve (worst agent).}
\end{figure}

\subsection{Discussion}
Our experimental results demonstrate similar trends to those shown in~\cite{Vinyals2017}. The variance observed in final performance achieved can be quite large, over different hyperparameter sets, different or same model parameter initializations and other stochasticity involved in learning. For Tables~\ref{tbl:average_reward} and~\ref{tbl:max_reward}, the higher the values the better. For Table~\ref{tbl:samples_required}, the lower the values the better. Among the 5 agents for BM, the best performing agent can achieve an average reward of 6.7 during testing, while the worst performing agent can barely achieve 0.1. Note that the average reward of 6.7 is twice more than the average reward of the best performing agent~(3) reported in~\cite{Vinyals2017} for BM. In addition, our method allows for an in-depth investigation into the agent's learning curves to identify which part of the learning was not effective and led to the sub-optimal final performance. We compare the best (average 6.7) and worst (average 0.1) agents based on their subgoal learning curves, and we find that the best agent is learning effectively across all subgoals. From Figure~\ref{fig:BM-curve-best}, the learning curves in all subtasks show consistent progress with more samples, where the learning curves of the worst agent show substantially less progress, often flat at zero with very rare spikes, as shown in Figure~\ref{fig:BM-curve-worst}. Especially for the \textit{BuildBarracks} subtask, the agent's learning is ineffective and it only occasionally stumbles upon the correct actions of building barracks at random and receives a corresponding reward signal. Alternatively, the comparison between the running average rewards for these two agents clearly demonstrates that learning for the best agent on the \textit{BuildBarracks} subtask is significantly more effective. The performance on this subtask also affects the final subtask \textit{BuildMarines} since without knowing how to build barracks, the agent cannot take the action of producing marines even if it has learnt this subpolicy. We believe such interpretability and explainability provided by our method are helpful in understanding and improving the learning process and the behavior of the agent.

On the other hand, the experimental results in CMAG show slightly less success. We believe this can be attributed to the difference in the setting of learning. In BM, the agent has to learn distinct skills and how to execute them in sequence in order to perform well, with relatively less emphasis on the degree of mastery of these skills. However, in CMAG the agent's mastery of the skills including mining minerals and gas directly and critically affects its final score, viz., total amount of minerals and gas collected. It means that the agent has to be able to perform the skills well, i.e., optimize with respect to time and manufacturing cost, which in itself can be a separate and more complex learning task. Another experimental difficulty for CMAG lies in the reward scales because the subtasks for collecting minerals and gas have high reward ceilings (as high as several thousand), while those for building the gas refineries have comparatively low reward ceilings (less than one hundred). Due to this large difference in the scales of the reward signals between subtasks, the learning on the subtasks is even more difficult and can be unbalanced.

\section{\label{sec:conclusion}Conclusion \& Future Work}
In this work, we examined the SC2 minigames and proposed a way to introduce human expertise to an HRL framework. By designing customized minigames to facilitate learning and leveraging the effectiveness of hierarchical structures in decomposing complex and large problems, we empirically showed that our approach is sample-efficient and enhances interpretability. This initial work invites several exploration directions: developing more efficient and effective ways of introducing human expertise; a more formal and principled state representation to further reduce the complexity of the state space (goal space) with theoretical analysis on its complexity; and a more efficient learning algorithm to pair with the HRL architecture, \textit{Experience Replay} and \textit{Curriculum Learning}.

\section{Acknowledgments}
This work was partially supported by an Academic Research Grant T1 251RES1827 from the Ministry of Education in Singapore and a grant from the Advanced Robotics Center at the National University of Singapore.

\bibliographystyle{aaai}
\bibliography{ICAPS_workshop.bib}

\begin{thebibliography}{}

\bibitem[\protect\citeauthoryear{Andrychowicz \bgroup et al\mbox.\egroup
  }{2017}]{HER-Andrychowicz2017}
Andrychowicz, M.; Wolski, F.; Ray, A.; Schneider, J.; Fong, R.; Welinder, P.;
  McGrew, B.; Tobin, J.; Pieter~Abbeel, O.; and Zaremba, W.
\newblock 2017.
\newblock Hindsight experience replay.
\newblock In Guyon, I.; Luxburg, U.~V.; Bengio, S.; Wallach, H.; Fergus, R.;
  Vishwanathan, S.; and Garnett, R., eds., {\em Advances in Neural Information
  Processing Systems 30}. Curran Associates, Inc.
\newblock  5048--5058.

\bibitem[\protect\citeauthoryear{Andrychowicz \bgroup et al\mbox.\egroup
  }{2020}]{robotics-OpenAI2018}
Andrychowicz, M.; Baker, B.; Chociej, M.; J{\'{o}}zefowicz, R.; McGrew, B.;
  Pachocki, J.; Petron, A.; Plappert, M.; Powell, G.; Ray, A.; Schneider, J.;
  Sidor, S.; Tobin, J.; Welinder, P.; Weng, L.; and Zaremba, W.
\newblock 2020.
\newblock {Learning dexterous in-hand manipulation}.
\newblock {\em International Journal of Robotics Research} 39(1):3--20.

\bibitem[\protect\citeauthoryear{Bakker and
  Schmidhuber}{2004}]{HRL-subgoalsubpolicy-Bakker2004}
Bakker, B., and Schmidhuber, J.
\newblock 2004.
\newblock Hierarchical reinforcement learning based on subgoal discovery and
  subpolicy specialization.
\newblock In {\em Proceedings of the 8-th Conference on Intelligent Autonomous
  Systems, IAS-8},  438--445.

\bibitem[\protect\citeauthoryear{Barto and
  Mahadevan}{2003}]{recent-hrl-Barto2003}
Barto, A.~G., and Mahadevan, S.
\newblock 2003.
\newblock Recent advances in hierarchical reinforcement learning.
\newblock {\em Discrete Event Dynamic Systems} 13(1–2):41–77.

\bibitem[\protect\citeauthoryear{Bengio \bgroup et al\mbox.\egroup
  }{2009}]{Bengio2009-curriculum-learning}
Bengio, Y.; Louradour, J.; Collobert, R.; and Weston, J.
\newblock 2009.
\newblock Curriculum learning.
\newblock In {\em Proceedings of the 26th Annual International Conference on
  Machine Learning}, ICML '09,  41–48.
\newblock New York, NY, USA: Association for Computing Machinery.

\bibitem[\protect\citeauthoryear{Buch, Ahmed, and
  Maruthappu}{2018}]{ai-medicine-Buch2018}
Buch, V.; Ahmed, I.; and Maruthappu, M.
\newblock 2018.
\newblock Artificial intelligence in medicine: Current trends and future
  possibilities.
\newblock {\em British Journal of General Practice} 68:143--144.

\bibitem[\protect\citeauthoryear{Cath}{2018}]{ai-law-Cath2018}
Cath, C.
\newblock 2018.
\newblock Governing artificial intelligence: Ethical, legal and technical
  opportunities and challenges.
\newblock {\em Philosophical Transactions of The Royal Society A Mathematical
  Physical and Engineering Sciences} 376:20180080.

\bibitem[\protect\citeauthoryear{Hauskrecht \bgroup et al\mbox.\egroup
  }{1998}]{macro-actions-Hauskrecht2013}
Hauskrecht, M.; Meuleau, N.; Kaelbling, L.~P.; Dean, T.; and Boutilier, C.
\newblock 1998.
\newblock Hierarchical solution of markov decision processes using
  macro-actions.
\newblock In {\em Proceedings of the Fourteenth Conference on Uncertainty in
  Artificial Intelligence}, UAI’98,  220–229.
\newblock San Francisco, CA, USA: Morgan Kaufmann Publishers Inc.

\bibitem[\protect\citeauthoryear{Hengst}{2010}]{hrl-springer-Hengst2010}
Hengst, B.
\newblock 2010.
\newblock Hierarchical reinforcement learning.
\newblock In Sammut, C., and Webb, G.~I., eds., {\em Encyclopedia of Machine
  Learning}. Boston, MA: Springer US.
\newblock  495--502.

\bibitem[\protect\citeauthoryear{Justesen \bgroup et al\mbox.\egroup
  }{2018}]{Justesen2018-illuminating-increasing-difficulty}
Justesen, N.; Torrado, R.~R.; Bontrager, P.; Khalifa, A.; Togelius, J.; and
  Risi, S.
\newblock 2018.
\newblock Illuminating generalization in deep reinforcement learning through
  procedural level generation.
\newblock In {\em NeurIPs Workshop on Deep Reinforcement Learning}.

\bibitem[\protect\citeauthoryear{Lee \bgroup et al\mbox.\egroup
  }{2018}]{modular-sc2-lee2018}
Lee, D.; Tang, H.; Zhang, J.~O.; Xu, H.; Darrell, T.; and Abbeel, P.
\newblock 2018.
\newblock Modular architecture for starcraft {II} with deep reinforcement
  learning.
\newblock In Rowe, J.~P., and Smith, G., eds., {\em Proceedings of the
  Fourteenth {AAAI} Conference on Artificial Intelligence and Interactive
  Digital Entertainment, {AIIDE} 2018, November 13-17, 2018, Edmonton, Canada},
   187--193.
\newblock {AAAI} Press.

\bibitem[\protect\citeauthoryear{Levy \bgroup et al\mbox.\egroup
  }{2019}]{HER-HRL-Evel2019}
Levy, A.; Konidaris, G.~D.; Jr., R.~P.; and Saenko, K.
\newblock 2019.
\newblock Learning multi-level hierarchies with hindsight.
\newblock In {\em 7th International Conference on Learning Representations,
  {ICLR} 2019, New Orleans, LA, USA, May 6-9, 2019}.

\bibitem[\protect\citeauthoryear{Li, Narayan, and
  Leong}{2017}]{CSRL-zhuoru2017-medcomp}
Li, Z.; Narayan, A.; and Leong, T.~Y.
\newblock 2017.
\newblock {An efficient approach to model-based hierarchical reinforcement
  learning}.
\newblock {\em 31st AAAI Conference on Artificial Intelligence, AAAI 2017}
  3583--3589.

\bibitem[\protect\citeauthoryear{Lillicrap \bgroup et al\mbox.\egroup
  }{2016}]{DDPG-Lillicrap2016}
Lillicrap, T.~P.; Hunt, J.~J.; Pritzel, A.; Heess, N.; Erez, T.; Tassa, Y.;
  Silver, D.; and Wierstra, D.
\newblock 2016.
\newblock {Continuous control with deep reinforcement learning}.
\newblock In {\em 4th International Conference on Learning Representations,
  ICLR 2016 - Conference Track Proceedings}.

\bibitem[\protect\citeauthoryear{Lin}{1993}]{experience-replay-Lin1993}
Lin, L.-J.
\newblock 1993.
\newblock {\em {Reinforcement learning for robots using neural networks}}.
\newblock Ph.D. Dissertation, Carnegie Mellon University.

\bibitem[\protect\citeauthoryear{Mnih \bgroup et al\mbox.\egroup
  }{2013}]{dqn-Mnih2013}
Mnih, V.; Kavukcuoglu, K.; Silver, D.; Graves, A.; Antonoglou, I.; Wierstra,
  D.; and Riedmiller, M.
\newblock 2013.
\newblock Playing atari with deep reinforcement learning.
\newblock In {\em NIPS Deep Learning Workshop}.

\bibitem[\protect\citeauthoryear{Mnih \bgroup et al\mbox.\egroup
  }{2015}]{human-level-control-Mnih2015}
Mnih, V.; Kavukcuoglu, K.; Silver, D.; Rusu, A.~A.; Veness, J.; Bellemare,
  M.~G.; Graves, A.; Riedmiller, M.~A.; Fidjeland, A.; Ostrovski, G.; Petersen,
  S.; Beattie, C.; Sadik, A.; Antonoglou, I.; King, H.; Kumaran, D.; Wierstra,
  D.; Legg, S.; and Hassabis, D.
\newblock 2015.
\newblock Human-level control through deep reinforcement learning.
\newblock {\em Nature} 518(7540):529--533.

\bibitem[\protect\citeauthoryear{Pang \bgroup et al\mbox.\egroup
  }{2019}]{SC2Fulllength-Pang2018}
Pang, Z.-J.; Liu, R.-Z.; Meng, Z.-Y.; Zhang, Y.; Yu, Y.; and Lu, T.
\newblock 2019.
\newblock {On Reinforcement Learning for Full-Length Game of StarCraft}.
\newblock In {\em Proceedings of the AAAI Conference on Artificial
  Intelligence}, volume~33,  4691--4698.

\bibitem[\protect\citeauthoryear{Ring}{2018}]{reaver}
Ring, R.
\newblock 2018.
\newblock Reaver: Modular deep reinforcement learning framework.
\newblock \url{https://github.com/inoryy/reaver}.

\bibitem[\protect\citeauthoryear{Schaul \bgroup et al\mbox.\egroup
  }{2015}]{uvfa-Schaul2015}
Schaul, T.; Horgan, D.; Gregor, K.; and Silver, D.
\newblock 2015.
\newblock Universal value function approximators.
\newblock In {\em Proceedings of the 32nd International Conference on
  International Conference on Machine Learning - Volume 37}, ICML’15,
  1312–1320.
\newblock JMLR.org.

\bibitem[\protect\citeauthoryear{Schulman \bgroup et al\mbox.\egroup
  }{2017}]{PPO-Schulman2017}
Schulman, J.; Wolski, F.; Dhariwal, P.; Radford, A.; and Klimov, O.
\newblock 2017.
\newblock Proximal policy optimization algorithms.
\newblock {\em CoRR} abs/1707.06347.

\bibitem[\protect\citeauthoryear{Silver \bgroup et al\mbox.\egroup
  }{2016}]{go-Silver2016a}
Silver, D.; Huang, A.; Maddison, C.~J.; Guez, A.; Sifre, L.; van~den Driessche,
  G.; Schrittwieser, J.; Antonoglou, I.; Panneershelvam, V.; Lanctot, M.;
  Dieleman, S.; Grewe, D.; Nham, J.; Kalchbrenner, N.; Sutskever, I.;
  Lillicrap, T.~P.; Leach, M.; Kavukcuoglu, K.; Graepel, T.; and Hassabis, D.
\newblock 2016.
\newblock Mastering the game of go with deep neural networks and tree search.
\newblock {\em Nature} 529:484--489.

\bibitem[\protect\citeauthoryear{Silver \bgroup et al\mbox.\egroup
  }{2017}]{alphaGo}
Silver, D.; Schrittwieser, J.; Simonyan, K.; Antonoglou, I.; Huang, A.; Guez,
  A.; Hubert, T.; Baker, L.; Lai, M.; Bolton, A.; Chen, Y.; Lillicrap, T.; Hui,
  F.; Sifre, L.; van~den Driessche, G.; Graepel, T.; and Hassabis, D.
\newblock 2017.
\newblock Mastering the game of go without human knowledge.
\newblock {\em Nature} 550(7676):354--359.

\bibitem[\protect\citeauthoryear{Singh \bgroup et al\mbox.\egroup
  }{2019}]{roboticRL}
Singh, A.; Yang, L.; Finn, C.; and Levine, S.
\newblock 2019.
\newblock End-to-end robotic reinforcement learning without reward engineering.
\newblock In Bicchi, A.; Kress{-}Gazit, H.; and Hutchinson, S., eds., {\em
  Robotics: Science and Systems XV, University of Freiburg, Freiburg im
  Breisgau, Germany, June 22-26, 2019}.

\bibitem[\protect\citeauthoryear{Sutton and Barto}{2018}]{RLbook2018}
Sutton, R.~S., and Barto, A.~G.
\newblock 2018.
\newblock {\em {Reinforcement Learning: An Introduction}}.
\newblock A Bradford Book55 Hayward Street Cambridge MA United States, second
  edition.

\bibitem[\protect\citeauthoryear{Sutton, Precup, and
  Singh}{1999}]{options-Sutton1999}
Sutton, R.~S.; Precup, D.; and Singh, S.
\newblock 1999.
\newblock Between mdps and semi-mdps: A framework for temporal abstraction in
  reinforcement learning.
\newblock {\em Artificial Intelligence} 112(1–2):181–211.

\bibitem[\protect\citeauthoryear{Vinyals \bgroup et al\mbox.\egroup
  }{2017}]{Vinyals2017}
Vinyals, O.; Ewalds, T.; Bartunov, S.; Georgiev, P.; Vezhnevets, A.~S.; Yeo,
  M.; Makhzani, A.; K{\"{u}}ttler, H.; Agapiou, J.~P.; Schrittwieser, J.; Quan,
  J.; Gaffney, S.; Petersen, S.; Simonyan, K.; Schaul, T.; van Hasselt, H.;
  Silver, D.; Lillicrap, T.~P.; Calderone, K.; Keet, P.; Brunasso, A.;
  Lawrence, D.; Ekermo, A.; Repp, J.; and Tsing, R.
\newblock 2017.
\newblock Starcraft {II:} {A} new challenge for reinforcement learning.
\newblock {\em CoRR} abs/1708.04782.

\bibitem[\protect\citeauthoryear{Vinyals \bgroup et al\mbox.\egroup
  }{2019}]{alphaStar}
Vinyals, O.; Babuschkin, I.; Czarnecki, W.~M.; Mathieu, M.; Dudzik, A.; Chung,
  J.; Choi, D.~H.; Powell, R.; Ewalds, T.; Georgiev, P.; Oh, J.; Horgan, D.;
  Kroiss, M.; Danihelka, I.; Huang, A.; Sifre, L.; Cai, T.; Agapiou, J.~P.;
  Jaderberg, M.; Vezhnevets, A.~S.; Leblond, R.; Pohlen, T.; Dalibard, V.;
  Budden, D.; Sulsky, Y.; Molloy, J.; Paine, T.~L.; Gulcehre, C.; Wang, Z.;
  Pfaff, T.; Wu, Y.; Ring, R.; Yogatama, D.; Wünsch, D.; McKinney, K.; Smith,
  O.; Schaul, T.; Lillicrap, T.; Kavukcuoglu, K.; Hassabis, D.; Apps, C.; and
  Silver, D.
\newblock 2019.
\newblock {Grandmaster level in StarCraft II using multi-agent reinforcement
  learning}.
\newblock {\em Nature} 575(7782):350--354.

\bibitem[\protect\citeauthoryear{Weng}{2020}]{weng2020curriculum-blog}
Weng, L.
\newblock 2020.
\newblock Curriculum for reinforcement learning.
\newblock {\em lilianweng.github.io/lil-log}.

\bibitem[\protect\citeauthoryear{Zambaldi \bgroup et al\mbox.\egroup
  }{2019}]{deep-relational-Zambaldi2019}
Zambaldi, V.~F.; Raposo, D.; Santoro, A.; Bapst, V.; Li, Y.; Babuschkin, I.;
  Tuyls, K.; Reichert, D.~P.; Lillicrap, T.~P.; Lockhart, E.; Shanahan, M.;
  Langston, V.; Pascanu, R.; Botvinick, M.; Vinyals, O.; and Battaglia, P.~W.
\newblock 2019.
\newblock Deep reinforcement learning with relational inductive biases.
\newblock In {\em 7th International Conference on Learning Representations,
  {ICLR} 2019, New Orleans, LA, USA, May 6-9, 2019}.

\bibitem[\protect\citeauthoryear{Zaremba and
  Sutskever}{2014}]{Zaremba2014-curriculum-task-specific}
Zaremba, W., and Sutskever, I.
\newblock 2014.
\newblock Learning to execute.
\newblock {\em CoRR} abs/1410.4615.

\end{thebibliography}

\end{document}